\documentclass[10pt,twocolumn,letterpaper]{article}

 \usepackage{cvpr}              
\usepackage{colortbl}
\usepackage{fancyhdr}

\usepackage[dvipsnames]{xcolor}

\definecolor{cvprblue}{rgb}{0.21,0.49,0.74}
\usepackage[pagebackref,breaklinks,colorlinks,citecolor=cvprblue]{hyperref}

\def\paperID{*****} 
\def\confName{CVPR}
\def\confYear{2024}

\title{T-DEED: Temporal-Discriminability Enhancer Encoder-Decoder for Precise Event Spotting in Sports Videos}

\author{Artur Xarles$^{1,2}$\hspace{0.7cm} Sergio Escalera$^{1,2,3}$\hspace{0.7cm} Thomas B. Moeslund$^{3}$\hspace{0.7cm} Albert Clapés$^{1,2}$ \\
$^{1}$Universitat de Barcelona, Barcelona, Spain\\
$^{2}$Computer Vision Center, Cerdanyola del Vallès, Spain \\
$^{3}$Aalborg University, Aalborg, Denmark\\
{\tt\small arturxe@gmail.com}, {\tt\small sescalera@ub.edu}, {\tt\small tbm@create.aau.dk}, {\tt\small aclapes@ub.edu} \\
}

\begin{document}
\maketitle

\begin{abstract}
In this paper, we introduce \textbf{T-DEED}, a Temporal-Discriminability Enhancer Encoder-Decoder for Precise Event Spotting in sports videos. T-DEED addresses multiple challenges in the task, including the need for discriminability among frame representations, high output temporal resolution to maintain prediction precision, and the necessity to capture information at different temporal scales to handle events with varying dynamics. It tackles these challenges through its specifically designed architecture, featuring an encoder-decoder for leveraging multiple temporal scales and achieving high output temporal resolution, along with temporal modules designed to increase token discriminability. Leveraging these characteristics, T-DEED achieves SOTA performance on the FigureSkating and FineDiving datasets. Code is available at \url{https://github.com/arturxe2/T-DEED}.
\end{abstract}
\vspace{-0.25cm}

\section{Introduction}
\label{sec:intro}

In recent years, remarkable progress has been made in the field of video understanding, largely driven by advancements in deep learning and increased computational power. These developments have enabled researchers to move beyond simpler tasks like action recognition~\cite{herath2017going} in trimmed videos to more complex challenges such as accurate localization of actions within untrimmed videos. Among these tasks, we find Temporal Action Localization (TAL), which represents actions as temporal intervals, and Action Spotting (AS), which uses single keyframes to represent them. While TAL~\cite{xia2020survey} has historically received more attention, AS has recently experienced an increase in research interest, particularly in domains such as sports, as fast-paced actions -- common in sports videos -- can be better represented by single temporal positions rather than intervals of time. Additionally, Hong et al.~\cite{hong2022spotting} extended AS to Precise Event Spotting (PES), distinguishing events from actions and employing tighter tolerances for evaluation. 

\begin{figure*}[t!]
  \includegraphics[width=\linewidth]{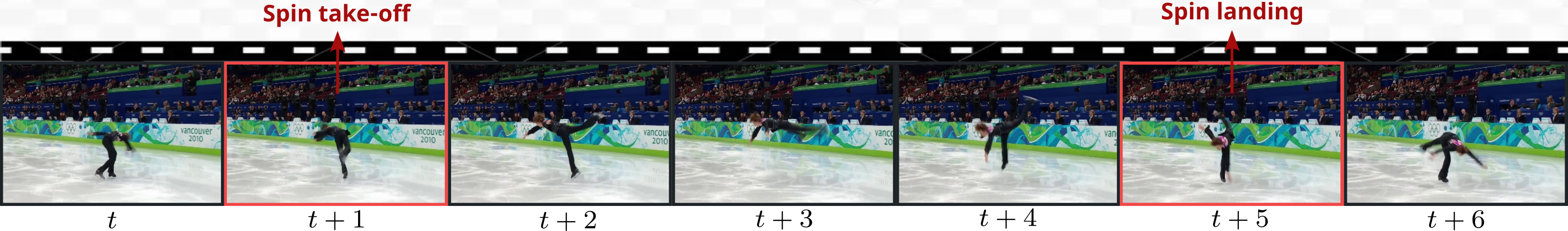}
  \caption{Illustration of the Precise Event Spotting task on the FigureSkating dataset. Red-marked frames contain events that require precise localization and correct classification among possible classes.}
  \label{fig:taskPES}
  \vspace{-0.15cm}
\end{figure*}

This paper specifically focuses on the task of Precise Event Spotting in sports (illustrated in Figure~\ref{fig:taskPES}), evaluated on FigureSkating~\cite{hong2021figureskatingdataset} and FineDiving~\cite{xu2022finediving}, two common sports datasets. Following Hong et al.~\cite{hong2022spotting}, we use tight tolerances to accommodate the fast-paced nature of sporting events, where even a minor temporal deviation of 1-2 frames can lead to missed events. Three main challenges faced by methods addressing this task include: (1) the need for discriminative per-frame representations to differentiate between adjacent frames with high spatial similarity, (2) the necessity of high output temporal resolution to avoid losing prediction precision, and (3) the variability in the amount of temporal context required for different events, influenced by dataset characteristics and event dynamics.

To tackle these challenges we introduce T-DEED, a Temporal-Discriminability Enhancer Encoder-Decoder for PES in sports videos. Introducing skip connections within its encoder-decoder architecture, T-DEED operates across various temporal scales, capturing actions requiring diverse temporal contexts, while restoring the original temporal resolution, thereby addressing challenges (2) and (3). Additionally, it integrates Scalable-Granularity Perception (SGP)~\cite{shi2023tridet} based layers to increase discriminability among features within the same temporal sequence, tackling challenge (1). To summarize, our main contributions are:

\begin{enumerate}
\item We incorporate residual connections into the SGP layer, enabling the integration of features from multiple temporal scales within the skip connections of our encoder-decoder architecture. This results in the creation of the SGP-Mixer layer, addressing challenge (3). 
\item We introduce the SGP-Mixer module within the SGP-Mixer layer, which adapts the SGP module to aggregate information from different temporal scales. This module shares the core principles of SGP to enhance token discriminability while modeling temporal information, therefore tackling challenge (1).
\item We conduct extensive ablations of T-DEED components, highlighting the advantages of the encoder-decoder architecture with SGP-based layers to enhance token discriminability. Additionally, T-DEED achieves state-of-the-art performance. Specifically, on FigureSkating, it improves mean Average Precision (mAP) by +1.15 for the FS-Comp split and +3.07 for the FS-Perf split. Similarly, on the FineDiving dataset, T-DEED achieves an improvement of +4.83 mAP. 
\end{enumerate}





\section{Related work}
\label{sec:related}

Over the past decade, deep learning has driven the field of video understanding through a remarkable evolution. Initially focused on simple tasks like classifying short-trimmed videos~\cite{abu2016youtube, carreira2017quo, goyal2017something}, the field has transitioned to more complex challenges, including Temporal Action Localization (TAL) and Action Spotting (AS). Both TAL and AS focus on temporally locating specific actions within untrimmed videos. TAL specifies temporal intervals for annotations, while AS represents actions with single keyframes, making TAL suitable for prolonged actions and AS more appropriate for fast-occurring actions, with the added benefit of reduced annotation costs. Furthermore, Hong et al.~\cite{hong2022spotting} extended the AS task to Precise Event Spotting (PES), introducing a key difference in the required precision of predictions, limited to only a few frames, and distinguishing between actions and events. In the sports domain, AS and PES are particularly popular as they adapt better to its fast nature, leading to the development of many approaches~\cite{hong2022spotting, soares2022temporally, xarles2023astra, cioppa2023soccernet, giancola2022soccernet, denize2024comedian} across different sports, including football~\cite{giancola2018soccernet, deliege2021soccernet}, figure skating~\cite{hong2021figureskatingdataset}, diving~\cite{xu2022finediving}, gymnastics~\cite{shao2020finegym}, and tennis~\cite{zhang2021tennisdataset}.

Given the inherent similarities between TAL and AS, the methodologies developed for these tasks frequently share common components. However, TAL methods have attracted more attention due to the earlier introduction of the task and a more extensive set of benchmarking datasets~\cite{idrees2017thumos, caba2015activitynet, zhao2019hacs, yeung2018every, damen2018scaling}, placing them a step ahead of AS methods. In contrast, AS methods have been tailored for specific datasets and competitive challenges, exemplified by their development in challenges like SoccerNet Action Spotting and SoccerNet Ball Action Spotting~\cite{giancola2022soccernet, cioppa2023soccernet}. Notably, E2E-Spot~\cite{hong2022spotting} stands out as the only action or event spotting method evaluated across multiple datasets.

\noindent{\textbf{Temporal Action Localization.}}
In TAL methods, a common classification divides them into two groups: two-stage methods~\cite{buch2017sst, escorcia2016daps, heilbron2016fast, qing2021temporal, xu2020g} and one-stage methods~\cite{lin2021learning, liu2022end, shi2023tridet, zhang2022actionformer, yang2022structured}. Two-stage methods generate class-agnostic proposals that are later classified into action labels or background, while one-stage models directly localize and classify actions in a single step, offering simplicity and achieving state-of-the-art performance in many TAL and AS scenarios. 

Among one-stage methods, early approaches~\cite{buch2019end, lin2017single} utilized anchor windows to generate action predictions. Later, Yang et al.~\cite{yang2020revisiting} introduced an anchor-free approach, relying on temporal points instead of anchor windows, highlighting the advantages of both techniques. Building on this approach, current state-of-the-art methods such as ActionFormer~\cite{zhang2022actionformer} and TriDet~\cite{shi2023tridet} have exhibited remarkable performance across various datasets. They leverage a feature pyramid network to process features at different temporal scales, a critical aspect for identifying actions that require distinct temporal contexts. The main difference lies in their prediction head and the layers used for feature processing. ActionFormer employs transformer layers, later revealed to suffer from the rank loss problem~\cite{dong2021attention}, negatively impacting token discriminability. To alleviate this problem, TriDet proposes a more efficient convolutional-based Scalable-Granularity Perception (SGP) layer, specially designed to increase token discriminability within the same temporal sequence, contributing to an improved overall performance. T-DEED, inspired by TAL trends, focuses on enhancing token discriminability while leveraging multiple temporal scales, with modifications tailored to meet the precision requirements of PES. 

\begin{figure*}[t]
  \centering
  \includegraphics[width=\linewidth]{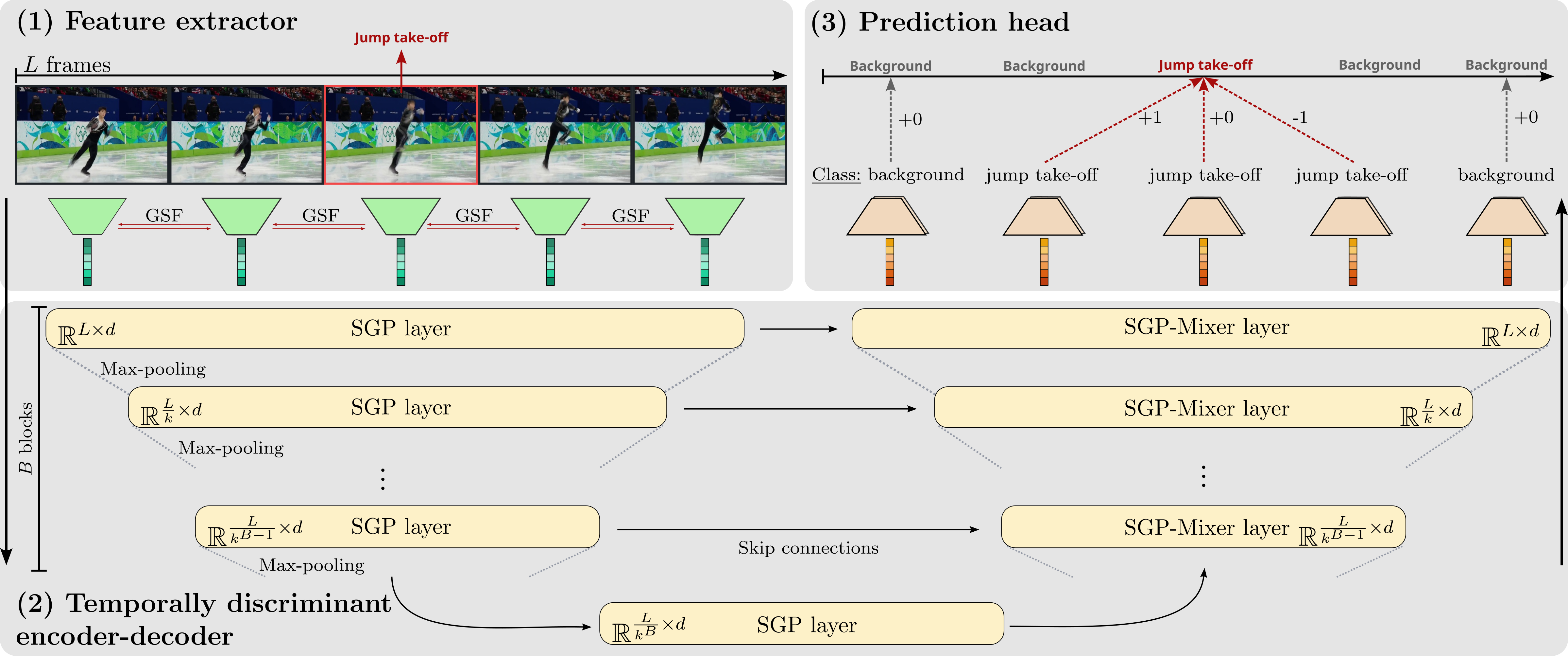}
  \caption{Illustration of \textbf{T-DEED} architecture comprising three key components: (1) \textbf{Feature extractor} to produce per-frame representations, (2) \textbf{Temporally discriminant encoder-decoder} to capture local and global temporal information while enhancing token discriminability, and (3) \textbf{Prediction head} to generate per-frame classifications and displacements for refinement.}
  \label{fig:modelarchitecture}
  \vspace{-0.25cm}
\end{figure*}

\noindent{\textbf{Action Spotting.}}
In AS, techniques similar to those in TAL have demonstrated state-of-the-art performance on the SoccerNet challenges~\cite{giancola2022soccernet, cioppa2023soccernet}. Many methods~\cite{soares2022temporally, xarles2023astra, denize2024comedian} classify temporal points as either background or actions and refine them through temporal regression using either convolutional~\cite{soares2022temporally} or Transformer-based~\cite{xarles2023astra, denize2024comedian} approaches. In contrast, Hong et al.~\cite{hong2022spotting} propose a simple end-to-end solution that employs a convolutional backbone with Gate-Shift Modules (GSM)~\cite{sudhakaran2020gate} for extracting per-frame features with short-term temporal information, followed by a Gated Recurrent Unit (GRU)~\cite{chung2014empirical} layer for long-term temporal information. This approach proves effective in their proposed task of PES across four different datasets~\cite{zhang2021tennisdataset, hong2021figureskatingdataset, shao2020finegym, xu2022finediving}, and is also adaptable to the coarser AS task in SoccerNet.

\noindent While many TAL and AS methods~\cite{soares2022temporally, xarles2023astra, shi2023tridet, zhang2022actionformer} rely on pre-extracted features due to their efficiency in training, end-to-end approaches have demonstrated that they can be beneficial in learning more meaningful features in some cases. This is exemplified by Hong et al.~\cite{hong2022spotting}, particularly in scenarios where precise predictions are essential, such as in the case of PES. Exploiting these advantages, T-DEED also adopts an end-to-end approach.


\section{Method}
\label{sec:methods}

\paragraph{\textbf{Problem definition.}} Precise Event Spotting (PES) involves the identification and localization of actions within an untrimmed video $X$, as illustrated in Figure~\ref{fig:taskPES}. Given the video input, the objective is to recognize and locate all the events occurring in the video, represented as $E = \{e_{1}, \dots, e_{N}\}$. The number of events, denoted as $N$, may vary across different videos. Each event instance $e_{i}$ comprises an action class $c_{i} \in \{1, \dots, C\}$ (where $C$ is the number of distinct event classes) and its corresponding temporal position $t_{i}$ (i.e. the exact frame where it occurs), forming a pair $e_{i} = (c_{i}, t_{i})$.

\paragraph{\textbf{Method overview.}} Our model, Temporal-Discriminability Enhancer Encoder-Decoder (T-DEED), is designed to increase token discriminability for Precise Event Spotting (PES) while leveraging multiple temporal scales. As illustrated in Figure~\ref{fig:modelarchitecture}, T-DEED comprises three main blocks: a feature extractor, a temporally discriminant encoder-decoder, and a prediction head.
We process videos through fixed-length clips, each containing $L$ densely sampled frames. The feature extractor, composed of a 2D backbone with Gate-Shift-Fuse (GSF) modules~\cite{sudhakaran2023gate}, handles the input frames, generating per-frame representations of dimension $d$, hereby referred to as \textit{tokens}. These tokens undergo further refinement within the temporally discriminant encoder-decoder. This module employs SGP layers which -- as shown by Shi et al.~\cite{shi2023tridet} -- diminish token similarity, thereby boosting discriminability across tokens of the same sequence.
The encoder-decoder architecture allows the processing of features across diverse temporal scales, helpful for detecting events requiring different amounts of temporal context. We also integrate skip connections to preserve the fine-grained information from the initial layers. To effectively merge information proceeding from varying temporal scales in the skip connections, we introduce the SGP-Mixer layer, detailed in Section~\ref{sec:methodsSGPMixer}. This layer employs the same principles as the SGP layer to enhance token discriminability while gathering information from varying range temporal windows. 
Finally, the output tokens are directed to the prediction head, resembling those commonly used in Action Spotting (AS) literature~\cite{xarles2023astra, soares2022temporally, denize2024comedian}. It encompasses a classification component to identify whether an event occurs at the given temporal position or in close proximity (within a radius of $r_{E}$ frames). Additionally, for the positive classifications, a displacement component pinpoints the exact temporal position of ground truth events.

Further details of the proposed method are discussed in the following sections. 

\subsection{Feature extractor}
The feature extractor processes the input frame sequence, $\mathbb{R}^{L\times H\times W \times 3}$ with $H\times W$ denoting the spatial resolution, and produces per-frame feature representations, $\mathbb{R}^{L\times d}$. Following Hong et al.~\cite{hong2022spotting}, we employ a compact 2D backbone to ensure efficiency and to accommodate longer video sequences. For comparability with previous PES methods, we choose RegNetY~\cite{radosavovic2020designing} as our feature extractor, known for its efficiency.
To incorporate local temporal information during the extractor, we utilize GSF~\cite{sudhakaran2023gate}, which overperforms GSM~\cite{sudhakaran2020gate} in action recognition tasks. GSF modules are specifically applied to the latter half of the RegNetY backbone, enabling local spatial modelling before integrating longer-term temporal context. 

\subsection{Temporally discriminant encoder-decoder}

The temporally discriminant encoder-decoder processes the tokens produced by the feature extractor module, which contain mainly spatial information, with the objective of enriching them with essential temporal information -- both local and global -- to enable precise event predictions. To accomplish this, it incorporates three key components: (1) an encoder-decoder architecture for exploiting information across diverse temporal scales, (2) SGP layers to increase discriminability among tokens within the same sequence, and (3) our novel SGP-Mixer layer, designed to effectively fuse features from varying temporal scales while maintaining the distinctiveness of its output tokens. \\

\noindent{\textbf{Encoder-decoder architecture.}}
Various works~\cite{zhang2022actionformer, shi2023tridet, soares2022temporally} have shown the benefits of processing features at different temporal scales in tasks related to action recognition. This is because certain actions inherently require longer temporal context for recognition and localization, while others can be identified with just a few frames. SOTA models in TAL achieve this by employing a feature pyramid, where the temporal dimension is downscaled by a factor of $k$ in the final layers through max-pooling, and predictions are made for each output feature at various temporal scales. However, in PES, where precision is crucial, this is detrimental. As shown in Section~\ref{sec:tridet}, the further down we go in the feature pyramid, the more deminishes the precision of the output predictions, inevitabely impacting model performance. To address this issue while still leveraging different temporal scales, we propose an encoder-decoder architecture. After temporal downscaling during the encoder, akin to the feature pyramid, we restore the original temporal resolution through the decoder, thereby regaining frame-level granularity for the representations. We do this by incorporating skip connections from the encoder to the decoder's upsampling. 

Specifically, the encoder begins by complementing the tokens with a learnable encoding that specifies its temporal position. Furthermore, it comprises $B$ encoder blocks, each consisting of an SGP layer to enhance discriminability while capturing temporal context, and a max-pooling operation to reduce the temporal dimension by a factor of $k$. An additional SGP layer is applied in the neck of the encoder-decoder, before passing the features to the decoder. In the decoder, the original temporal resolution is restored through $B$ decoder blocks. Each of these blocks incorporates an SGP-Mixer layer, which extends the SGP layer to increase the temporal dimension by a factor of $k$ while integrating information coming from the skip connections. \\

\begin{figure}[tb]
  \centering
  \includegraphics[width=\linewidth]{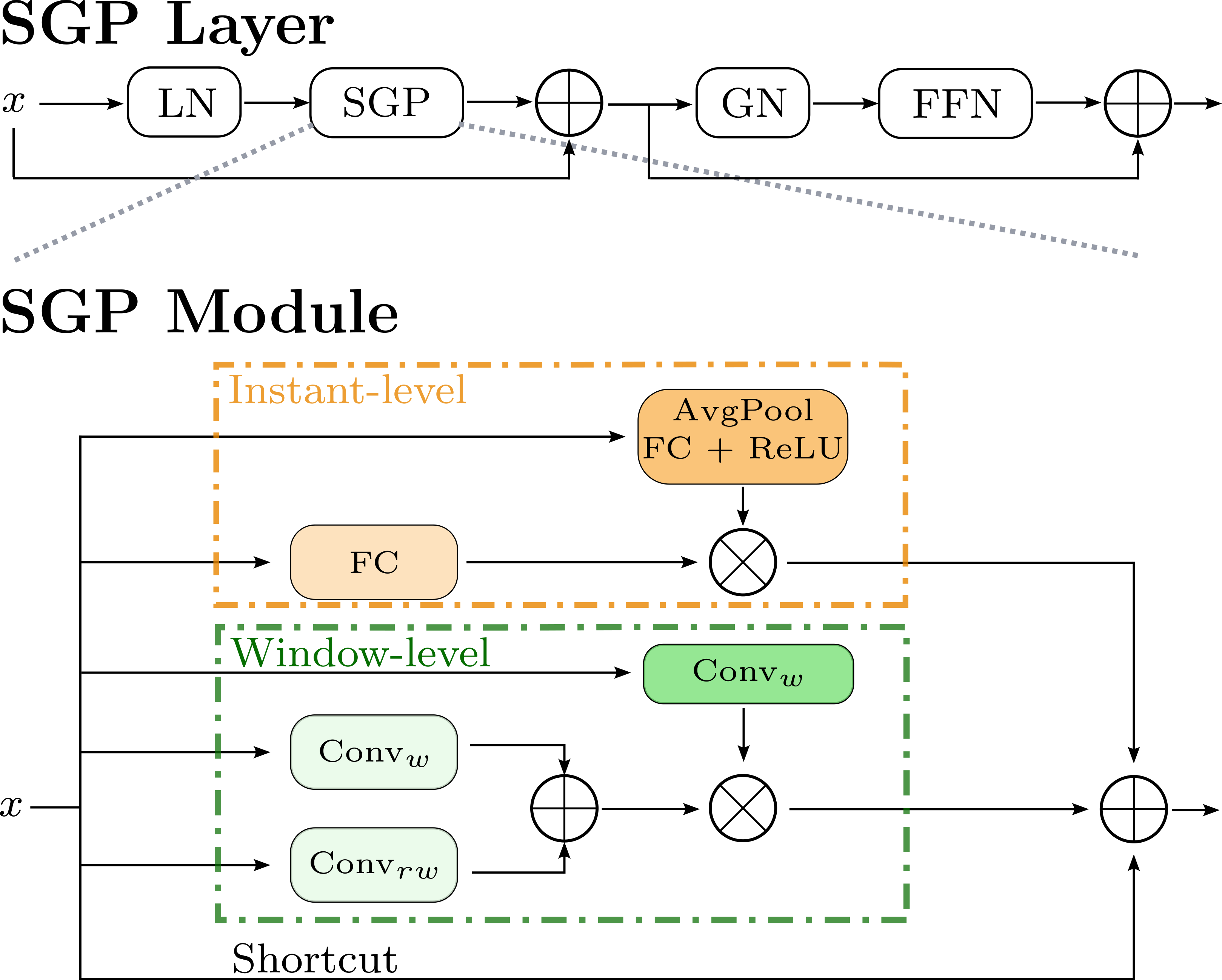}
  \caption{Illustration of the SGP layer structure, with its module comprising an instant-level branch to boost token discriminability and a window-level branch for temporal modeling.}
  \label{fig:sgplayer}
  \vspace{-0.25cm}
\end{figure}

\noindent{\textbf{SGP layer.}}
The SGP layer, as introduced by Shi et al.~\cite{shi2023tridet}, addresses the rank-loss problem~\cite{dong2021attention} commonly encountered in Transformers~\cite{vaswani2017attention}, thus improving token discriminability within sequences. As depicted in the top of Figure~\ref{fig:sgplayer}, it replaces the self-attention module in the Transformer layer with an SGP module illustrated at the bottom of the figure. This module comprises two primary branches: an instant-level branch and a window-level branch. The instant-level branch aims to boost token discriminability by increasing its distance from the clip-level average token, while the window-level branch captures temporal information from multiple receptive fields. Furthermore, the SGP layer substitutes one of the layer normalization modules with group normalization. 

Given its advantageous characteristics, which contributed to Tridet's SOTA results in TAL, we believe it can be even more beneficial in PES, where precise frame-level predictions are crucial and can benefit from improved discriminability between concurrent tokens, as discussed in Section~\ref{sec:ablationsSGP}. \\

\begin{figure}[tb]
  \centering
  \includegraphics[width=\linewidth]{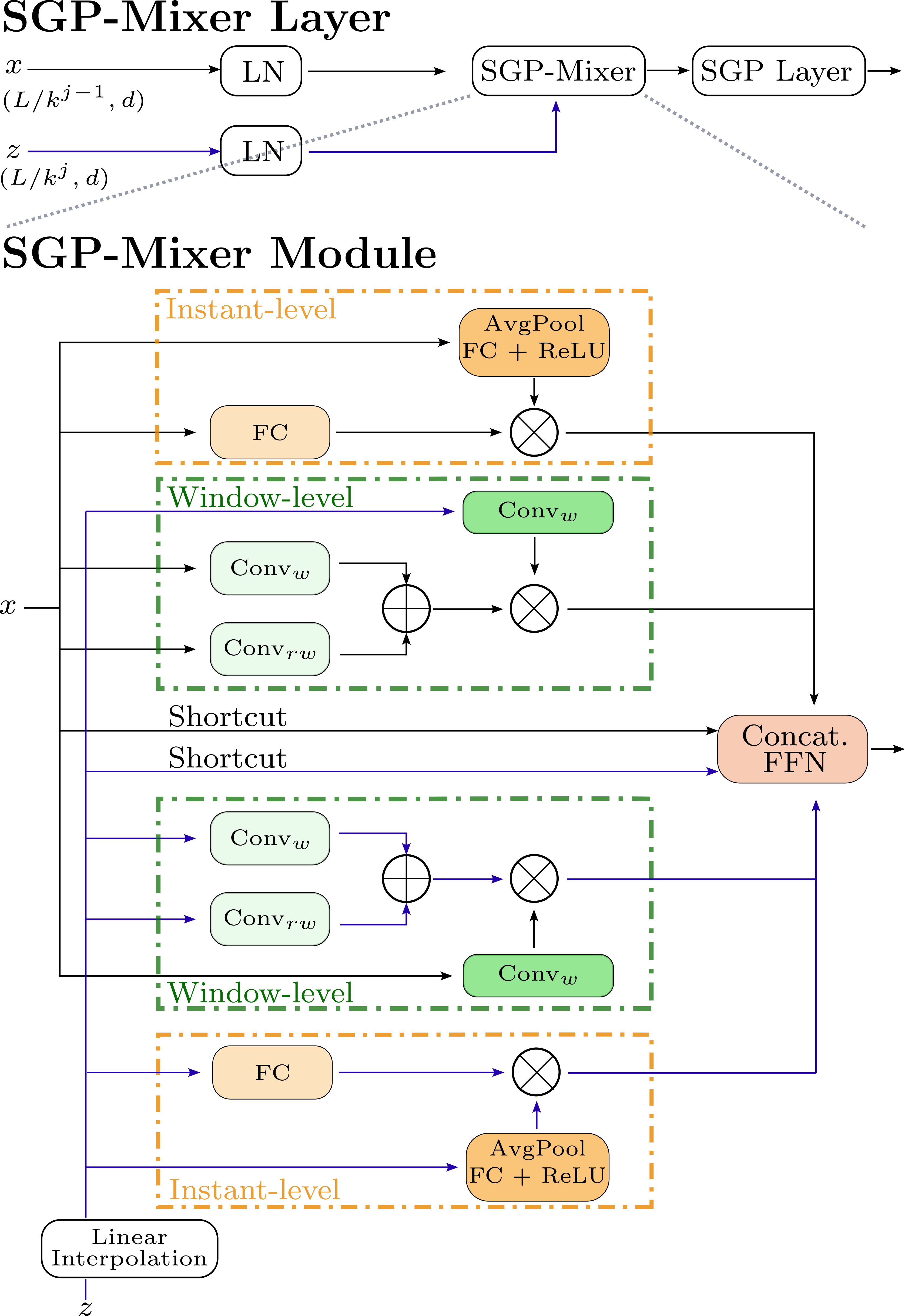}
  \caption{Illustration of the SGP-Mixer layer structure integrating an SGP-Mixer module to aggregate features of different temporal scales. This module follows the SGP principles, incorporating instant-level and window-level branches to boost token discriminability while merging the features.}
  \label{fig:sgpmixerlayer}
\end{figure}

\noindent{\textbf{SGP-Mixer layer.}}
\label{sec:methodsSGPMixer}
As shown in the top of Figure~\ref{fig:sgpmixerlayer}, the SGP-Mixer layer extends the original SGP layer to accommodate two inputs with distinct temporal scales. This adaptation becomes necessary in our architecture due to the reception of an input feature $z\in \mathbb{R}^{L/k^{j}\times d}$ from the preceding layer, and features $x\in \mathbb{R}^{L/k^{(j-1)}\times d}$ from the skip connection, where $j\in \{1, \dots, B\}$ denotes the depth within the decoder blocks. Both features undergo layer normalization before entering the SGP-Mixer module, where a combination of both features is generated. Finally, the output features are further processed through identical components within an SGP layer.

The fusion of both features is done in the SGP-Mixer module, shown at the bottom of the Figure~\ref{fig:sgpmixerlayer}. First, it upsamples the features from the previous layer to match the temporal dimensions ($\mathbb{R}^{L/k^{j}} \rightarrow \mathbb{R}^{L/k^{(j-1)}})$ using linear interpolation. Following the principles of the SGP layer, it has two instant-level branches -- one for each input feature -- to improve token discriminability. This is particularly important for the features in the previous layer, since upsampling may produce similar tokens in adjacent temporal positions. In addition, it includes two window-level branches to aggregate information from both features while capturing different temporal contexts. In each branch, one feature evolves to mix information across different temporal receptive fields, which is later gated by the other feature. Shortcuts are also introduced, but, unlike SGP, feature aggregation from different branches involves concatenation and linear projection, which we find more effective than simple addition, as will be shown later in Section~\ref{sec:ablationsED}. 

\subsection{Prediction head}
\label{sec:methodsPH}
Following common AS approaches~\cite{xarles2023astra, soares2022temporally, denize2024comedian}, we include a prediction head consisting of a classification head and a displacement head. The classification head uses a linear layer and a softmax activation to project the output of the temporally discriminant encoder-decoder, $\mathbb{R}^{L \times d}$, onto $\hat{y}^{c}\in\mathbb{R}^{L \times (C+1)}$, representing the probability of each temporal position containing each of the events or a background class. Similarly, the displacement head uses a linear layer to project the same output to $\hat{y}^{d}\in\mathbb{R}^{L \times 1}$, representing the displacement toward the ground truth event if an actual event is present at the corresponding temporal position. 

\subsection{Training details}

The model is trained using a combination of a classification loss ($\mathcal{L}_{c}$) and a displacement loss ($\mathcal{L}_{d})$. For classification, per-frame cross-entropy loss is employed weighting the positive classes by a factor of $w$ for all events. Mean squared error is used for displacement. The final loss ($\mathcal{L}$) for a given clip is the sum of both losses:
\vspace{-0.25cm}
\begin{equation}
\label{eq:lossC}
\mathcal{L} = \mathcal{L}_c + \mathcal{L}_d = \frac{1}{L} \sum_{l=1}^{L} CE_w(y^{c}_{l}, \hat{y}^{c}_{l}) + MSE(y^{d}_{l}, \hat{y}^{d}_{l}),
\end{equation}
where $y^{c}_{l} $ represents the one-hot encoding of the event in frame $l$, $\hat{y}^{c}_{l}$ denotes the classification probabilities at that frame, $y^{d}_{l}$ indicates the actual displacement to a ground-truth event (if an event is within the detection radius $r_{E}$), and $\hat{y}^{d}_{l}$ represents the predicted displacement.

During training, we perform typical data augmentation techniques such as random cropping, random horizontal flip, gaussian blur, color jitter, and mixup~\cite{zhang2017mixup}. 

\subsection{Inference}
At inference time, the data augmentation techniques are disabled and we use clips with 50\% of overlapping. Moreover, to reduce the number of candidate events, Soft Non-Maximum Suppression~\cite{bodla2017soft} is applied. 

\begin{table*}[t]
  \caption{Comparison with SOTA methods on FigureSkating (FS-Comp and FS-Perf splits) and FineDiving datasets, highlighting the best results in bold and the second-top performance with underlining. For each model, we report feature extractor sizes in MegaFlops (MF) and input modalities including RGB images and Optical Flow (OF).}
  \label{tab:sota}
  \centering
  \begin{tabular}{lcccccccc}
    \toprule
   & & & \multicolumn{2}{c}{FS-Comp}  & \multicolumn{2}{c}{FS-Perf} & \multicolumn{2}{c}{FineDiving}  \\
     Model & Size & Input & $\delta$ = 1 & 2 & 1 & 2 & 1 & 2 \\
    \midrule
    E2E-Spot~\cite{hong2022spotting} & 200MF & RGB & 81.00 & \underline{93.50} & 85.10 & 95.70 & 68.40 & 85.30 \\
    E2E-Spot~\cite{hong2022spotting} & 800MF & RGB & 84.00 & - & 83.60 & - & 64.60 & - \\
    \rowcolor{lightgray}
    E2E-Spot~\cite{hong2022spotting} & 800MF & RGB + OF~\cite{horn1981determining} & 83.40 & \textbf{94.90} & 83.30 & \underline{96.00} & 66.40 & 84.80 \\
    \hline
    T-DEED & 200MF & RGB & \textbf{85.15} & 91.70 & \underline{86.79} & \textbf{96.05} & \underline{71.48} & \underline{87.62}  \\
    T-DEED & 800MF & RGB & \underline{84.77} & 92.86 & \textbf{88.17} & 95.87 & \textbf{73.23} & \textbf{88.88} \\
    \bottomrule
  \end{tabular}
  \vspace{-0.25cm}
\end{table*}

\newpage
\section{Results}
\label{sec:results}

In this section, we detail the evaluation setup for our proposed method and present experimental results demonstrating its superiority over current SOTA methods in PES on the application domain of sports data. Additionally, we conduct ablations on key components of our proposal.

\subsection{Datasets}
\label{sec:datasets}
We conduct experiments on two challenging datasets: FigureSkating~\cite{hong2021figureskatingdataset} and FineDiving~\cite{xu2022finediving}. FigureSkating comprises 11 videos featuring 371 short-program performances from Winter Olympics and World Championships, annotated with 4 event classes. Annotations cover $0.23\%$ of the frames. Evaluation is performed on two different splits: FS-Comp and FS-Perf, as detailed in Hong et al.~\cite{hong2022spotting}. FineDiving~\cite{xu2022finediving} consists of 3000 diving clips annotated with 4 different event classes, covering $2.2\%$ of the frames. Both datasets' videos have a frame rate between 25 and 30 frames per second. 

\subsection{Evaluation}
Following standard practices, we train on the train split, use the validation split for early stopping, and evaluate on the test split for all datasets. Model performance is assessed using the mAP metric, which calculates the mean of Average Precisions across different events. We employ a tight version of the metric with a tolerance of $\delta=1$ frame and a loose metric with $\delta=2$ frames. 

\subsection{Implementation details}

We train T-DEED using clips of $L=100$ frames with a batch size of 8 clips. Each epoch comprises 5000 clips, randomly sampled from the training videos. Models are trained for 50 epochs using AdamW optimizer~\cite{loshchilov2017decoupled}, with a base learning rate of 8e-04, with 3 linear warmup epochs, and cosine decay. Positive classes in the cross-entropy loss are weighted with $w=5$. We evaluate two versions of our feature extractor, denoted as RegNetY-200MF and RegNetY-800MF with hidden dimensions set to $d=368$ and $d=768$, respectively, and a downscaling factor of $k=2$. Additional dataset-sensitive model hyperparameters are outlined in the supplementary material.

\subsection{Comparison to SOTA}
In Table~\ref{tab:sota}, we compare our proposed T-DEED in two variations, with smaller and larger feature extractors, against previous SOTA models across four different configurations and datasets. We present mAP scores with $\delta=1$ and $\delta=2$. For the Figure Skating dataset, in both FS-Comp and FS-Perf configurations, our models exhibit improvements on the tight metric, with gains of up to +1.15 and +3.07, respectively. However, performance on the loose metric, with a 2-frame tolerance, is comparable for FS-Comp and slightly worse in FS-Perf. We attribute this to the use of dilation in the E2E-Spot model for the FigureSkating dataset, which may be beneficial when using larger tolerances. Additionally, in the FineDiving dataset, T-DEED clearly outperforms the SOTA on both metrics, with improvements of up to +4.87 on the tight metric and +3.58 on the loose metric.

\subsection{Ablations}
In this section, we incrementally ablate the different components of our approach. We start with a base model consisting of the feature extractor without GSF, producing features with only spatial information, and plain layers of the different temporal approaches. Initially, we consider only the classification head (i.e., $\mathcal{L} = \mathcal{L}_c$), resulting in per-frame classifications. For each approach, we evaluate multiple configurations with different hyperparameters (e.g., number of layers) and present results with the best configuration, reporting the mAP with a tolerance of $\delta=1$. Each experiment is trained with two different seeds and we report the average for robustness. Ablations are conducted using two datasets: FineDiving and FigureSkating on the FS-Comp split. 

\begin{table}[t]
  \caption{Ablation of T-DEED's main components using mAP with $\delta=1$, highlighting the best results in bold. 
  }
  \label{tab:ablation_unet}
  \centering
  \begin{tabular}{llcc}
    \toprule
    \multicolumn{2}{c}{} & FS-Comp & FineDiving \\
    \multicolumn{2}{l}{Experiments} & mAP ($\delta=1$) & mAP ($\delta=1$) \\
    \midrule
    \multicolumn{2}{l}{\textit{(a) Temporal module}} & & \\
     & Transformer & 69.68 & 59.84 \\
    & GRU & 72.04 & 59.09 \\
    & SGP & \textbf{73.14} & \textbf{60.07} \\
    \midrule
    \multicolumn{2}{l}{\textit{(b) Skip connection}} &  &  \\
    & w/o & 74.29 & 63.01 \\
    & sum & 73.27 & 60.14 \\
    & concat & 76.78 & 61.90 \\
    & SGP-Mixer (sum) & 76.88 & 61.34 \\
    & SGP-Mixer & \textbf{77.96} & \textbf{63.67} \\
    \bottomrule
\end{tabular}
\label{tab:main_ablations}
\vspace{-0.25cm}
\end{table}

\noindent{\textbf{Enhancing token discriminability.}}
\label{sec:ablationsSGP}
As previously discussed, tasks like PES require tokens with sufficient discriminability to prevent precision loss in predictions due to similar representations for spatially similar adjacent frames. Here, we evaluate the discriminability of three commonly used layers for modeling temporal information: Transformer~\cite{vaswani2017attention}, GRU~\cite{chung2014empirical}, and SGP~\cite{shi2023tridet}. We measure the discriminability of their output tokens by averaging the cosine similarity between each token and the mean token of the sequence, as described in Shi et al.~\cite{shi2023tridet}.

\begin{figure}[t]
\centering
\begin{subfigure}[t]{1\linewidth}
\centering
   \includegraphics[width=1\linewidth]{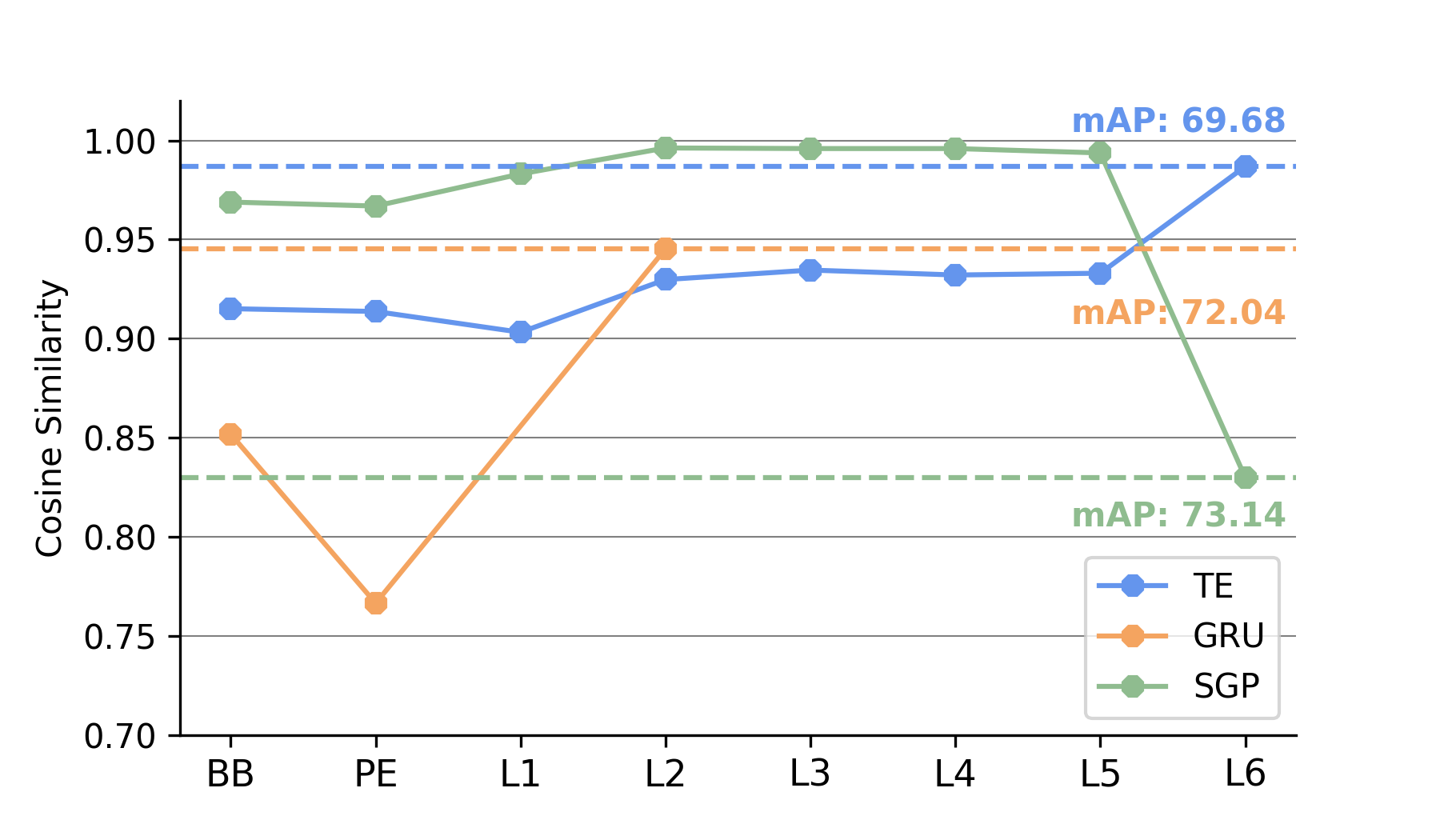}
   \caption{FS-Comp discriminability analysis.}
   \label{fig:discriminabilityFS} 
\end{subfigure}

\begin{subfigure}[b]{1\linewidth}
\centering
   \includegraphics[width=1\linewidth]{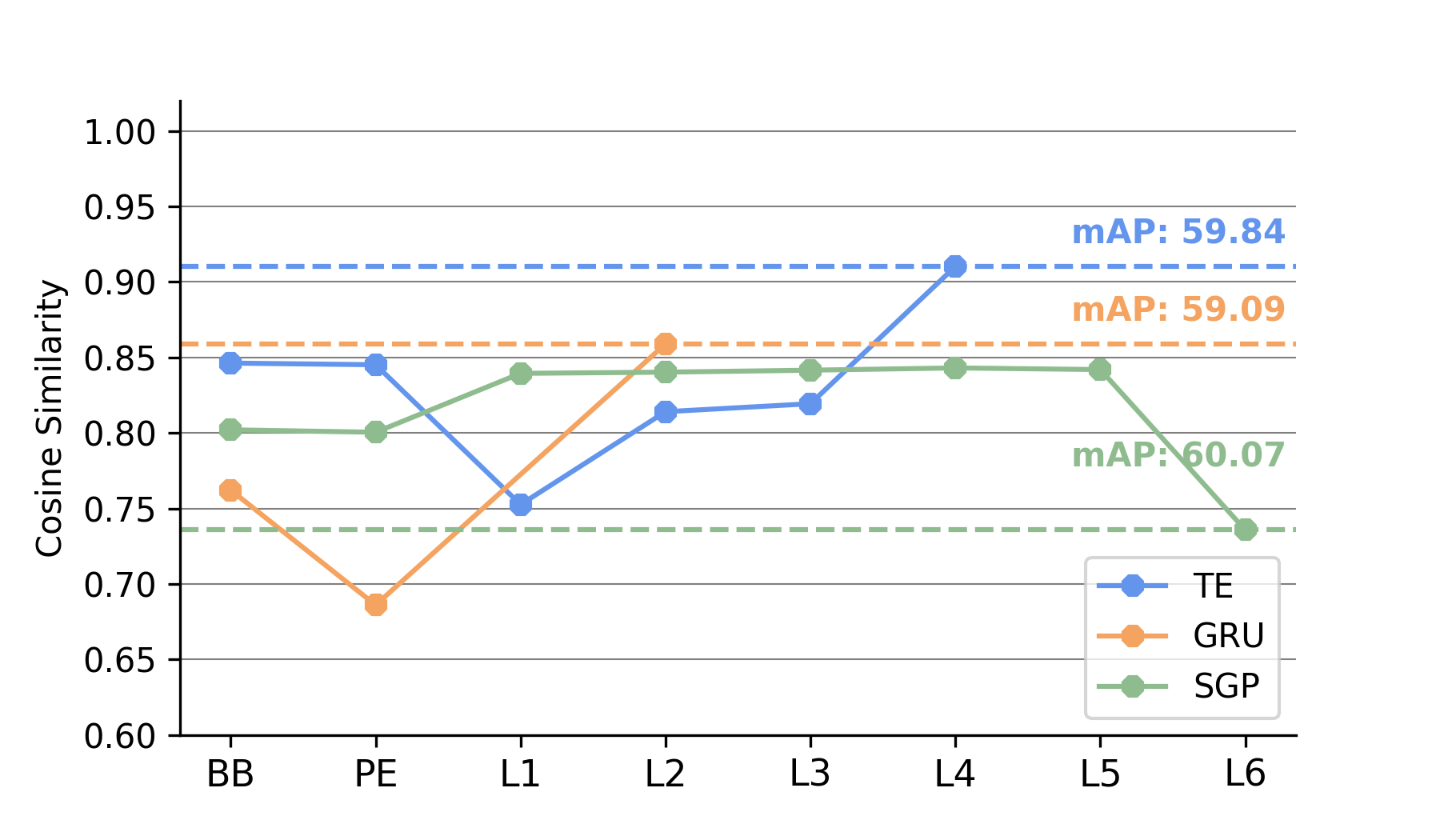}
   \caption{FineDiving discriminability analysis.}
   \label{fig:discriminabilityFD}
\end{subfigure}
\caption{\textbf{Temporal module discriminability analysis.} Cosine similarity after backbone (BB), post-positional encoding (PE), and at each temporal layer is displayed. Additionally, mAP performance with $\delta=1$ is reported.}
\label{fig:discriminability}
\vspace{-0.45cm}
\end{figure}

Figure~\ref{fig:discriminability} illustrates the rank-loss problem in Transformers, where token similarity increases across layers, therefore losing discriminability. In contrast, GRU layers, typically requiring fewer layers, show slight improvement in similarity. However, SGP layers exhibit the lowest similarity (i.e., highest discriminability) among output tokens, demonstrating its effectiveness in enhancing token discriminability. Additionally, this enhanced discriminability leads to the best performance among the different temporal modules, as shown in Table~\ref{tab:main_ablations}a. This observation highlights the advantages of employing the SGP layer, particularly in precision-demanding tasks like PES.

\noindent{\textbf{Defining multiple temporal scales.}}
\label{sec:ablationsED}
Employing multiple temporal scales while processing videos has proven effective across various TAL and AS methodologies~\cite{shi2023tridet, zhang2022actionformer, soares2022temporally}. Here, we evaluate our encoder-decoder architecture, operating on multiple temporal scales, using diverse mixture approaches in skip connections. Specifically, we assess five variations: (1) no skip connections, (2) addition, (3) concatenation and linear projection, (4) a modified SGP-Mixer aggregating branches information with summation, and (5) our proposed SGP-Mixer layer. Further details of each approach can be found in the supplementary material.

Results of these approaches are presented in Table~\ref{tab:main_ablations}b, where we observe that all encoder-decoder-based approaches, with or without skip connections, outperform the baseline using a single scale. This underscores the importance of capturing information from multiple temporal scales, effectively achieved in our encoder-decoder. However, further analysis reveals issues with approaches utilizing addition for feature aggregation within skip connections, as they tend to yield inferior results compared to alternative methods. We hypothesize that this problem may be due to the attribution of equal weight to features from skip connections and previous layers when adding them. This could be particularly critical in the top layers of the architecture, where information passed from skip connections may be too primitive. Finally, our proposed SGP-Mixer layer stands as the best approach for both datasets in aggregating information within the skip connections. This emphasizes the advantages of our carefully designed SGP-Mixer layer in aggregating information across multiple temporal scales while enhancing token discriminability. 

\begin{table}[t]
  \caption{Further ablations and analysis of T-DEED using mAP with $\delta=1$. 
  }
  \label{tab:ablation_unet}
  \centering
  \begin{tabular}{llcc}
    \toprule
    \multicolumn{2}{c}{} & FS-Comp & FineDiving \\
    \multicolumn{2}{l}{Experiments} & mAP ($\delta=1$) & mAP ($\delta=1$) \\
    \midrule
    \multicolumn{2}{l}{\textit{(a) Displacement head}} & & \\
    & w/o \& dilation = 0 & 74.74 & 63.67 \\
    & w/o \& dilation = 1 & 77.96 & 61.97 \\
    & $r_{E}$ = 1 & \textbf{78.20} & 67.49 \\
    & $r_{E}$ = 2 & 77.12 & \textbf{68.31} \\
    \midrule
    \multicolumn{2}{l}{\textit{(b) Feature pyramids}} & & \\
    & Tridet & 68.31 & 64.28 \\
    \midrule
    \multicolumn{2}{l}{\textit{(c) Feature extractor}} & & \\
    & w/ gsm & 81.00 & 67.95 \\
    & w/ gsm (half) & 81.05 & 67.47 \\
    & w/ gsf & 80.43 & 67.87 \\
    & w/ gsf (half) & \textbf{81.25} & \textbf{68.40} \\
    \midrule
    \multicolumn{2}{l}{\textit{(d) Clip length}} & & \\
    & $L=25$ frames & 71.02 & 66.61 \\
    & $L=50$ frames & 76.87 & 64.84 \\
    & $L=100$ frames & \textbf{81.25} & \textbf{68.40} \\
    & $L=200$ frames & 79.19 & 65.29 \\
    \midrule
    \multicolumn{2}{l}{\textit{(e) Postprocessing}} & & \\
    & NMS~\cite{neubeck2006efficient} & 81.27 & 68.40 \\
    & SNMS~\cite{bodla2017soft} & \textbf{85.15} & \textbf{71.48}\\
    \bottomrule
    
\end{tabular}
\label{tab:ablations}
\vspace{-0.25cm}
\end{table}

\noindent{\textbf{Introducing the displacement head.}}
In Table~\ref{tab:ablations}a we evaluate various methods for addressing the imbalance between frames containing actual events and background frames in PES. These methods include label dilation, which aims to detect events within a radius around the ground truth but may impact prediction precision, and the utilization of a displacement head, as detailed in Section ~\ref{sec:methodsPH}. Results indicate that the prediction head offers more benefits compared to label dilation, enabling wider range of event detection without sacrificing prediction precision. 

\begin{figure}[tb]
  \centering
  \includegraphics[width=\linewidth]{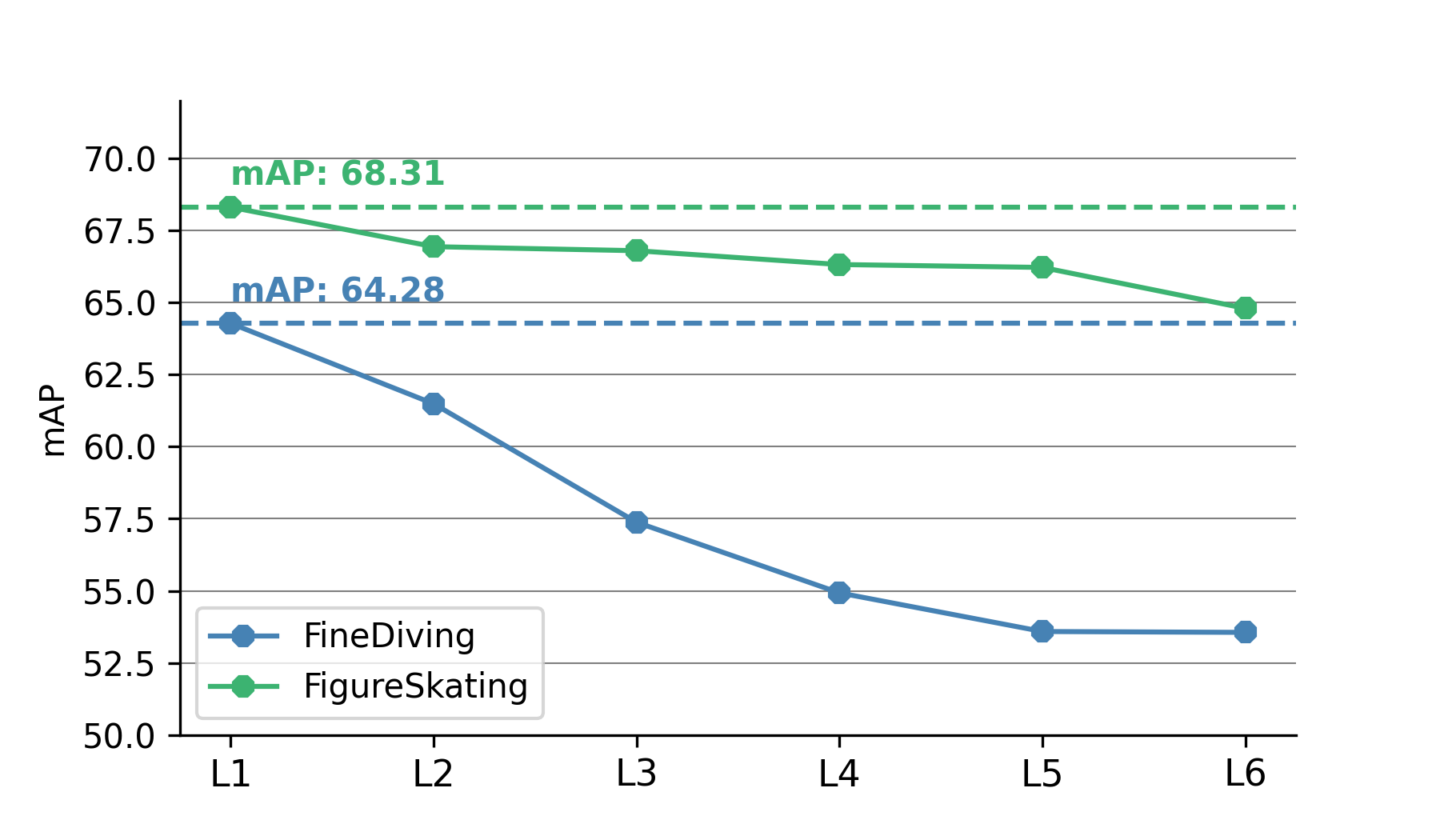}
  \caption{Per-layer mAP analysis using the Feature Pyramid Network. mAP is reported at each layer of the SGP feature pyramid, accumulating predictions from previous layers, for the FineDiving and FigureSkating datasets.}
  \label{fig:tridet}
  \vspace{-0.25cm}
\end{figure}

\noindent{\textbf{Feature pyramids.}}
\label{sec:tridet}
A common approach in TAL to process multiple temporal scales is the use of feature pyramids, which resembles using only the encoder of our approach. In Table~\ref{tab:ablations}b, we explore adapting the SGP Feature Pyramid proposed in Shi et al.~\cite{shi2023tridet} to our task. This method generates predictions at each temporal scale and integrates a displacement head to locate them. However, we observe a performance decrease compared to our encoder-decoder architecture. Further analysis of the predictions produced at each layer of the pyramid, depicted in Figure~\ref{fig:tridet}, reveals a decrease in performance as we descend the pyramid towards more high-level but lower-resolution features. This suggests that generating predictions at lower resolutions might compromise the accuracy. Our encoder-decoder architecture is able to model high-level information while recovering the original temporal resolution, thus avoiding losing the temporal precision that is critical in PES. 


\noindent{\textbf{Feature extractor ablations.}}
In Table~\ref{tab:ablations}c, we explore the impact of integrating different temporal shift modules into our 2D backbone to capture local temporal context. Specifically, we consider two modules, GSM~\cite{sudhakaran2020gate} and GSF~\cite{sudhakaran2023gate}, known for their effectiveness in action recognition tasks. We assess two variants: one applying these modules across all backbone layers, and another limited to the latter half to promote stronger spatial modeling before temporal integration. Following ~\cite{hong2022spotting}, we apply these modules to $\frac{1}{4}$ of the channels of the residual blocks. Results indicate that incorporating local temporal modeling into the backbone is especially useful in the FigureSkating dataset, while FineDiving shows comparable results to those without temporal modules. Notably, GSF applied to the latter half of the backbone yields the best performance for both datasets.

\noindent{\textbf{Clip length analysis.}}
Regarding optimal clip length, Table~\ref{tab:ablations}d evaluates various number of frames. In FigureSkating, performance improves with increasing clip length, plateauing at 100 frames. Similarly, results for FineDiving also indicate 100 frames as the optimal choice.

\noindent{\textbf{Postprocessing analysis.}}
In Table~\ref{tab:ablations}e, we illustrate the impact of employing two postprocessing techniques: Non-Maximum Suppression (NMS)~\cite{neubeck2006efficient} and Soft Non-Maximum Suppression (SNMS)~\cite{bodla2017soft}. The optimal results are achieved with a 1-frame window for NMS and a 3-frame window for SNMS. Notably, SNMS consistently improves mAP for both datasets.

\section{Conclusion}
\label{sec:conclusion}
This work presented T-DEED, a model designed to address Precise Event Spotting across various sports datasets. Ablation studies underscore the importance of processing videos in multiple temporal scales and enhancing token discriminability for precise predictions. To address these challenges, we integrate an encoder-decoder architecture and propose the SGP-Mixer layer, aimed at aggregating information at various temporal scales within skip connections while improving token discriminability. Additionally, T-DEED achieves SOTA performance on the FigureSkating and FineDiving datasets.\\

\noindent {\textbf{Acknowledgements.}} This work has been partially supported by the Spanish project PID2022-136436NB-I00 and by ICREA under the ICREA Academia programme.

{
    \small
    \bibliographystyle{ieeenat_fullname}
    \bibliography{main}
}

\newpage

\maketitlesupplementary
\appendix

\section{Implementation Details for T-DEED}

Here we outline the configuration used for each T-DEED model in the SOTA comparison from Table 1 in the main paper. 

All models utilize data augmentations, including mixup with $\alpha=\beta=0.2$, color jitter with probability $0.25$, and Gaussian blur with probability $0.25$. For FigureSkating, frames of size $398\times 224$ are randomly cropped to $224\times 224$, while for FineDiving and FineGym, frames are resized to $224\times 224$. The detection radius $r_{E}$ is set to 2 frames for FineDiving, and 1 frame for FigureSkating. Additionally, we weight the positive classes with $w=5$ within the cross-entropy loss. 

Among model-specific hyperparameters, we have the number of blocks ($B$), the kernel size ($ks$), and the scalable factor within the SGP module ($r$). These are chosen independently for FineDiving, FS-Comp, and FS-Perf: 
\begin{itemize}
    \item FineDiving (T-DEED w/ 200MF): $B=2$, $ks=7$, $r=4$.
    \item FineDiving (T-DEED w/ 800MF): $B=2$, $ks=9$, $r=4$.
    \item FS-Comps(T-DEED w/ 200MF): $B=3$, $ks=5$, $r=2$.
    \item FS-Comp (T-DEED w/ 800MF): $B=2$, $ks=9$, $r=4$.
    \item FS-Perf (T-DEED w/ 200MF): $B=3$, $ks=9$, $r=2$.
    \item FS-Perf (T-DEED w/ 800MF): $B=2$, $ks=9$, $r=4$.
\end{itemize}

\section{Decoder blocks alternative approaches}

\begin{figure}[h]
\centering
\begin{subfigure}[t]{1\linewidth}
\centering
   \includegraphics[width=0.98\linewidth]{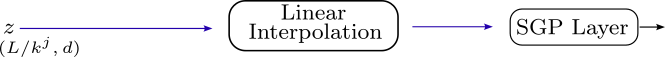}
   \caption{Decoder block without skip connections}
   \label{fig:noskip} 
\end{subfigure}
\vspace{0.15cm}
\medskip
\hrulefill\par
\begin{subfigure}[b]{1\linewidth}
\centering
   \includegraphics[width=0.98\linewidth]{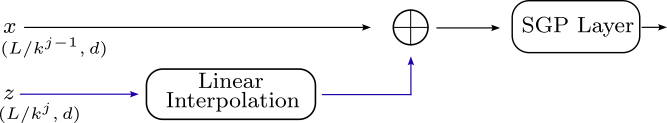}
   \caption{Decoder block with addition on the skip connections.}
   \label{fig:skipsum}
\end{subfigure}
\vspace{0.15cm}
\medskip
\hrulefill\par
\begin{subfigure}[b]{1\linewidth}
\centering
   \includegraphics[width=0.98\linewidth]{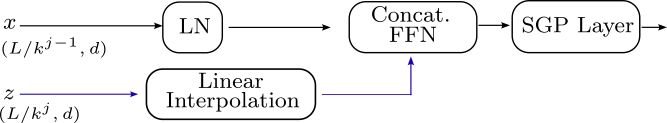}
   \caption{Decoder block with concatenation on the skip connections.}
   \label{fig:skipconcat}
\end{subfigure}
\vspace{0.15cm}
\medskip
\hrulefill\par
\begin{subfigure}[b]{1\linewidth}
\centering
   \includegraphics[width=0.98\linewidth]{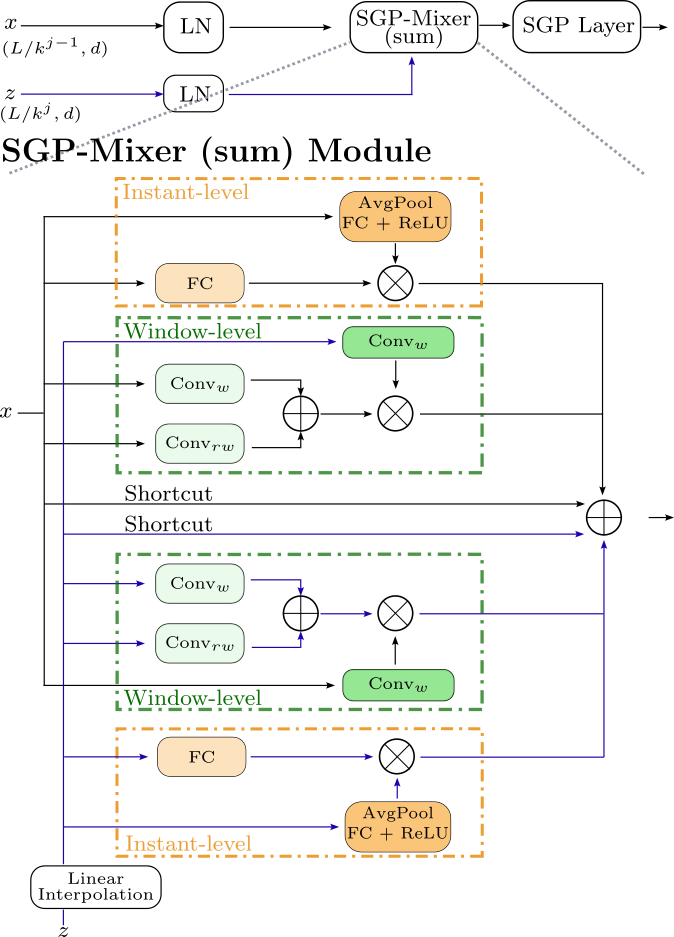}
   \caption{Decoder block with SGP-Mixer merging branches information with addition.}
   \label{fig:sgpmixersum}
\end{subfigure}
\caption{Decoder block alternative approaches.}
\label{fig:skips}
\end{figure}

In Figure~\ref{fig:skips}, we illustrate different approaches to the decoder block design and the information aggregation within skip connections. Figure~\ref{fig:noskip} depicts the case without skip connections, where information from the previous layer undergoes linear interpolation and an SGP Layer processes it. Figure~\ref{fig:skipsum} illustrates aggregation through addition, where upsampled features from the previous layer are added to skip connection features before further processing with an SGP Layer. Similarly, Figure~\ref{fig:skipconcat} shows concatenation-based aggregation, where concatenated features undergo linear projection before SGP Layer processing. Finally, Figure~\ref{fig:sgpmixersum} showcases an approach akin to our SGP-Mixer layer but using addition instead of concatenation and linear projection for feature aggregation. 

\end{document}